\documentclass[10pt,twocolumn,letterpaper]{article} 

\usepackage{avss}
\usepackage{times}
\usepackage{epsfig}
\usepackage{graphicx}
\usepackage{amsmath}
\usepackage{amssymb}
\usepackage{tipa}
\usepackage{svg}
\usepackage{booktabs}
\usepackage{subfig}

\newcommand\blfootnote[1]{%
  \begingroup
  \renewcommand\thefootnote{}\footnote{#1}%
  \addtocounter{footnote}{-1}%
  \endgroup
}

\PassOptionsToPackage{hyphens}{url}\usepackage[pagebackref=true,breaklinks=true,letterpaper=true,colorlinks,bookmarks=false]{hyperref}

\newif\ifnotes
\notestrue

\avssfinalcopy 

\def\avssPaperID{102} 

\newcommand{\baseline}{Baseline-50\xspace}

\ifavssfinal\fi
\begin{document}

\title{Human Pose Estimation for Real-World Crowded Scenarios}
\author{
Thomas Golda\textsuperscript{1,2,*}\quad
Tobias Kalb\textsuperscript{2,*}\quad
Arne Schumann\textsuperscript{2}\quad
J\"urgen Beyerer\textsuperscript{1,2}\\
\begin{minipage}[t]{0.5\linewidth}
\vspace{.05cm}
\begin{center}
\textsuperscript{1}Vision and Fusion Lab\\
Karlsruhe Institute of Technology KIT\\
c/o Technologiefabrik, Haid-und-Neu-Strasse 7\\
76131 Karlsruhe, Germany
\end{center}
\end{minipage}
\begin{minipage}[t]{0.5\linewidth}
\vspace{.05cm}
\begin{center}
\textsuperscript{2}Fraunhofer Institute for Optronics, System\\Technologies and Image Exploitation IOSB\\
Fraunhoferstrasse 1\\
76131 Karlsruhe, Germany
\end{center}
\vspace{.05cm}
\end{minipage}\\
{\tt\small firstname.lastname@iosb.fraunhofer.de}\\
}

\maketitle
\begin{abstract}
Human pose estimation has recently made significant progress with the adoption of deep convolutional neural networks and many applications have attracted tremendous interest in recent years.
However, many of these applications require pose estimation for human crowds, which still is a rarely addressed problem.
For this purpose this work explores methods to optimize pose estimation for human crowds, focusing on challenges introduced with larger scale crowds like people in close proximity to each other, mutual occlusions, and partial visibility of people due to the environment.
In order to address these challenges, multiple approaches are evaluated including: the explicit detection of occluded body parts, a data augmentation method to generate occlusions and the use of the synthetic generated dataset JTA~\cite{fabbri2018learning}.
In order to overcome the transfer gap of JTA originating from a low pose variety and less dense crowds, an extension dataset is created to ease the use for real-world applications.
\end{abstract}

\section{Introduction}
\begin{figure}[t]
    \begin{center}
        \includegraphics[width=\linewidth]{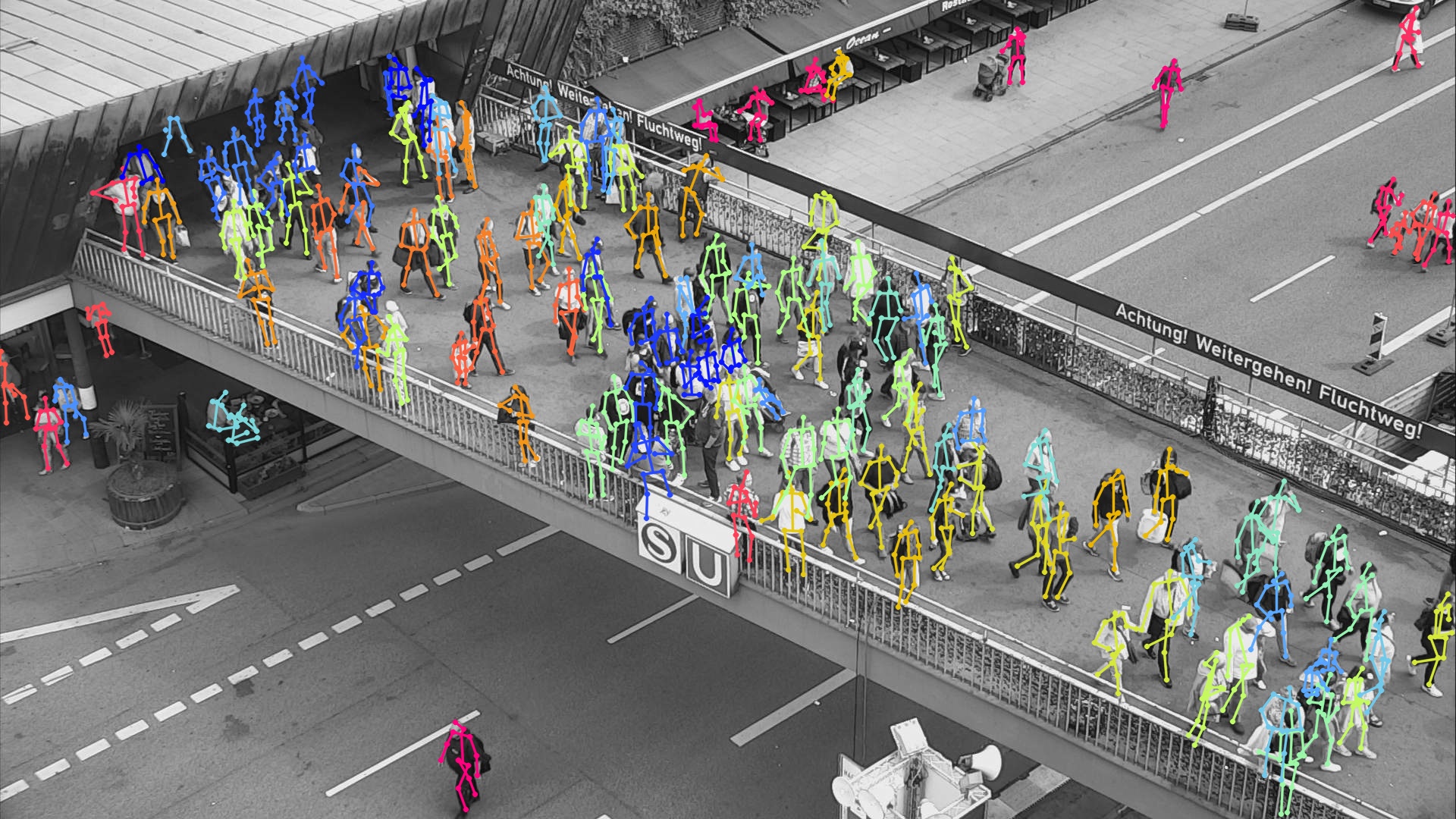}
    \end{center}
    \caption{\textbf{Example of a scene captured by a surveillance camera.} Such situation is characterized by heterogeneous levels of crowdedness and lots of occlusions and ambiguities at more crowded spots. The drawn poses show the difficulty of the task.}
    \label{fig:crowdedscene}
\end{figure}

Articulated human pose estimation has been a commonly addressed problem in computer vision for many years.
It has attracted broad interest for its many applications such as video surveillance, action recognition, human computer interaction and motion capturing.
Recent advances in human pose estimation have been achieved by harnessing the power of deep convolutional neural networks (CNNs) and large-scale pose estimation datasets like COCO~\cite{LinMBHPRDZ14} and MPII~\cite{andriluka14cvpr}.
Up until now most approaches focus on pose estimation for images with few people in uncrowded scenarios.
\blfootnote{\textsuperscript{*} Both authors contributed equally to this work.}
However, in many applications it is inevitable to handle larger groups or crowds of people which also introduces many new challenges, see Fig.~\ref{fig:crowdedscene}. 
Such challenges include partially occluded people, mutual occlusions by humans and low resolution of persons due to sensor limitations or distance to the camera.
One prominent application is video based surveillance of public places, where pose estimation can be utilized to analyze crowd behavior. 
To be more clear at this point, our notion of crowded situations is different compared to the field of e.g. crowd counting.
Whilst in the latter case most of the time only the area around the head of single persons is visible, we regard highly crowded situations as cases in which occlusions between persons are not permanently present but still dominate the situation.
Pose information encoded as body joint keypoints is a low dimensional description and hence well suited for real-time analysis.
This could enable to detect and prevent critical or dangerous situations by alarming security forces, if necessary.
However, to the best of our knowledge there exists very little work on human pose estimation for crowded surveillance scenarios.

Thus, in this work we evaluate several aspects of state-of-the-art pose detection methods with respect to the characteristics of crowded surveillance imagery.
Specifically, we focus on three distinct aspects in the pipeline of modern pose detection methods.
We first evaluate the impact of a recently proposed occlusion data augmentation method on highly crowded data.
We then propose our own architecture modification for explicitly modeling occluded keypoints within a CNN pose detector. 
And finally, we generate a new simulated dataset, which is specifically suited to evaluate recent pose detection methods under crowded surveillance scenario conditions.
Our resulting surveillance pose recognition model, as well as the generated dataset will be made available to the community to spark further research.
\\
This work is divided as follows.
Starting with this introduction, this work then presents related literature for human pose estimation as well as the few existing works on human pose estimation for crowds.
The subsequent section concentrates on the methods chosen for our experiments, wWihich includes data augmentation methods, the adaption of an enhanced architecture for detection of occluded keypoints and the use of synthetic training data.
Finally, the described methods are evaluated separately and in a combined model.
\section{Related Work}
\subsection{Top-Down and Bottom-Up Methods}
Existing methods for human pose estimation can be divided into two opposing approach classes: bottom-up and top-down methods.
Methods of the former class attempt to first locate all person keypoints (e.g. joints, eyes, ears) over the entire input image.
These keypoints are then grouped to single persons, which is done by solving an assignment problem.
In contrast, top-down methods start from person detections and identify keypoints within the detection bounding boxes, following the assumption that each bounding box contains at most one keypoint per keypoint class.
Each of them comes with different advantages and disadvantages.
Although bottom-up methods seem to be better suited for crowded scenes since they process whole images and their runtime is thus less dependent of the actual number of persons,  literature shows that top-down methods like \cite{li2018crowdpose} and \cite{xiao2018simple} perform comparably well on publicly available~datasets.

\subsection{Simple Baselines for Human Pose Estimation}
The model introduced in \cite{xiao2018simple} is the recent winner of the PoseTrack ECCV 2018\footnote{\url{https://posetrack.net/workshops/eccv2018/posetrack_eccv_2018_results.html}} challenge in the category for multi-person pose tracking. 
It provides a simplistic baseline model based on a top-down pose estimation network and a novel method for tracking detected poses across multiple frames.
The main difference to other state-of-the-art methods like CPN~\cite{pyramidnetwork} and the hourglass network \cite{stackedhourglass} lies in the method for generating high resolution feature maps using transposed~convolutions.\\
The method proposed by \cite{xiao2018simple} refines the common top-down pipeline by additionally using a second set of bounding boxes, which is determined by optical flow from previous frames. 
Therefore, the keypoints in the current frame are propagated to the next frame using the optical flow between the frames. 
With the help of the propagated keypoints, a bounding box proposal is generated.
Non-maximum suppression is then used to select the best set of bounding boxes.
Furthermore, the authors apply a flow-based pose similarity measure $S_{\text{flow}}$ to the greedy matching algorithm, instead of using Intersection-over-Union~(IoU). 

\subsection{Further Work on Crowd Pose Estimation}
As previously stated, human pose estimation in crowded scenarios is a rarely addressed problem.
However, some recent works focus on this subject.
Li~\etal~\cite{li2018crowdpose} introduce a new benchmark for evaluating pose estimation methods for this problem and proposed a method where human bounding box proposals obtained by human detector are fed into joint-candidate single person pose estimator (JCSPPE). 
JCSPPE locates the joint candidates with different response scores on the heatmap. 
Then the joint association algorithm takes these results and builds a person-joint connection graph. 
Finally, a graph matching problem is solved to find the best joint association result with a global maximum joints association algorithm.
With their method the authors show robustness to interference in crowded scenes. 
Yet, their definition of "crowdedness" differs slightly from common situations in surveillance applications.
Since the collected data originates from the COCO dataset, it has a relatively low person per image count.
Furthermore, the perspective the images were taken from often differs strongly from typical views of surveillance cameras, which in most cases are mounted in higher spots leading to a steeper viewing~angle.\\
Furthermore, the work of Fabbri~\etal~\cite{fabbri2018learning} is also of great interest for this kind of problem.
Using \emph{Grand~Theft~Auto~V}, a widely known computer game with an active modding community, they created a synthetic dataset of humans annotated with highly accurate keypoint information alongside with a tool that allows to easily generate own synthetic datasets. 
Both, \cite{fabbri2018learning} and \cite{li2018crowdpose} proposed interesting datasets, which we used for our experiments, whereas \cite{fabbri2018learning} is more similar to surveillance situations than the dataset introduced~by~\cite{li2018crowdpose}.
\section{Methods}
This section explores methods to optimize pose estimation for crowd applications. 
For this purpose we adapt the single-person pose estimator proposed by Xiao \etal~\cite{xiao2018simple} using the ResNet50~\cite{7780459} network as backbone.
The method is a two-staged top-down approach that first localizes every person using state-of-the-art object detection methods and then performs single-person pose estimation for every detection. 
This choice is motivated by the simplicity of the model and its performance compared to other state-of-the-art approaches. 

For adaptation to crowded scenes, data augmentation methods are first discussed which improve the performance of the baseline model on both uncrowded and crowded scenarios. 
Next, the baseline architecture is extended in order to explicitly detect occluded keypoints with the help of the occluded keypoint annotation provided with the JTA dataset. 
Finally, the earlier mentioned JTA dataset is examined for the challenges of real world crowd applications and extended by a small dataset which was created specifically to alleviate some of these~problems.
\subsection{Synthetic Generated Occlusions}
As S{\'a}r{\'a}ndi \etal~\cite{Sarandi18IROSW} have shown, synthetic generated occlusions are an effective augmentation method to reduce occlusion-induced errors in pose estimation, while also improving results on data without any occlusions. 
Since the amount of real-world data that is currently available for pose estimation of human crowds is very limited, the question arises whether these augmentation methods are also helpful in the task of pose estimation for human crowds. 
In order to answer this question, two different kinds of synthetic occlusions are examined: occlusions by random objects and occlusions by person instances. 
For both augmentation methods the instance segmentation labels provided with the COCO dataset are used to get a high variety of objects and persons.\\
Similar to the approach of S{\'a}r{\'a}ndi \etal~\cite{Sarandi18IROSW} randomly picked objects from the COCO dataset are cut out from their original images using the provided segmentation maps. These cutouts are then added onto the bounding boxes of people in the training image. The size and the position of the cutout is chosen at random, where the size can range from  8\% to 70\% of the size of the person's bounding box. 
Furthermore, the occlusion flag of a keypoint is adjusted accordingly, if it is occluded by the cutout.\\
While the segmentation labels for person instances are also available, they need to be handled separately, because they introduce different challenges than the occlusion by objects.
For example, if a person is added exactly on top of another person, the labeled keypoints are likely to be occluded by the added person's body which can lead to faulty training data. 
To counterbalance this issue, two different methods are evaluated to add additional persons to the detections.
The first one is instead of adding full person instances to the image only body parts are added onto the scene.
Secondly, placement of cutouts is restricted to the outer bounds of the target persons bounding box to avoid confusing the top-down pose recognizer with two fully valid person images inside one bounding box.

\subsection{Occlusion Detection Networks}
\begin{figure}
    \begin{center}
        \includegraphics[width=\linewidth]{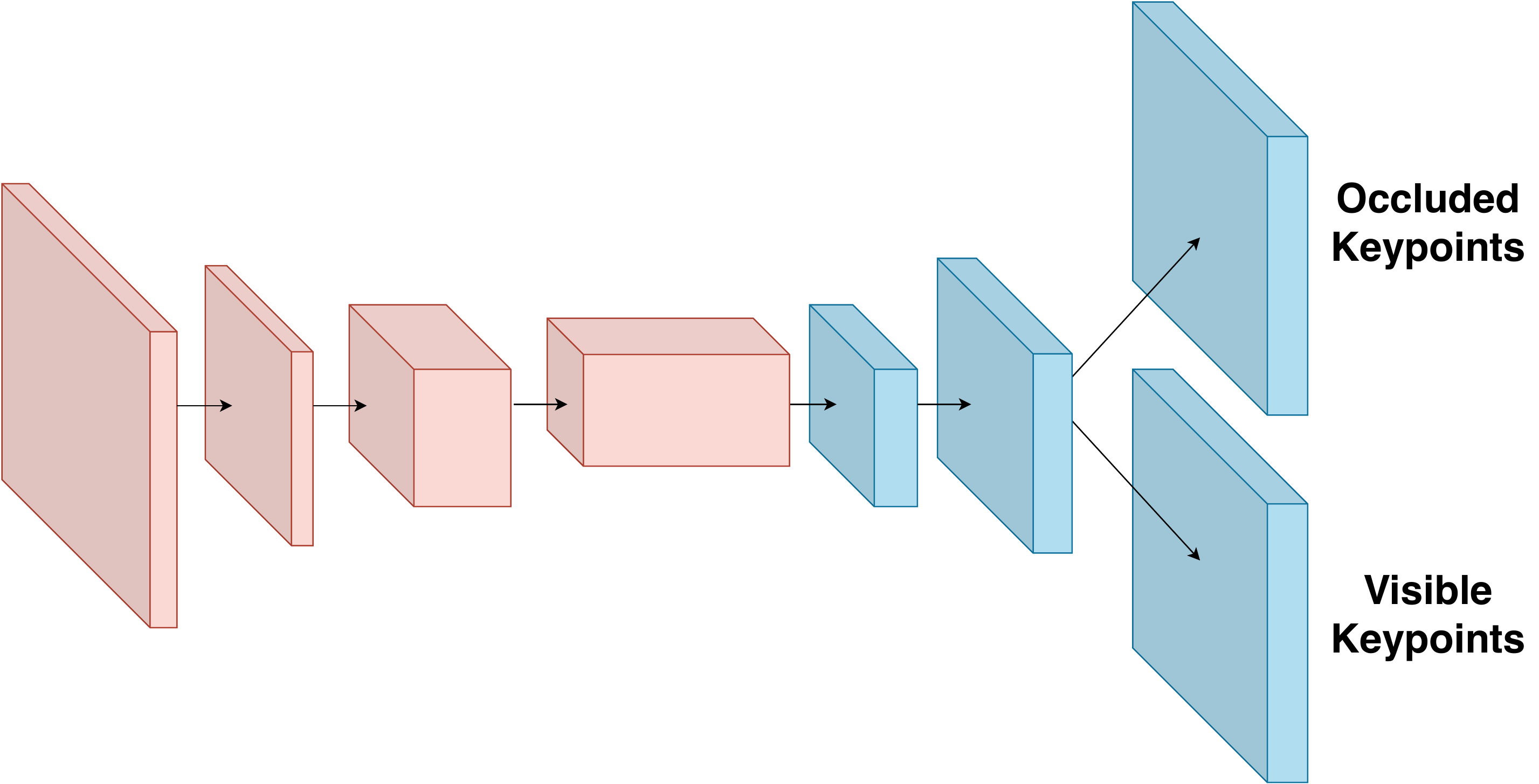}
        \includegraphics[width=\linewidth]{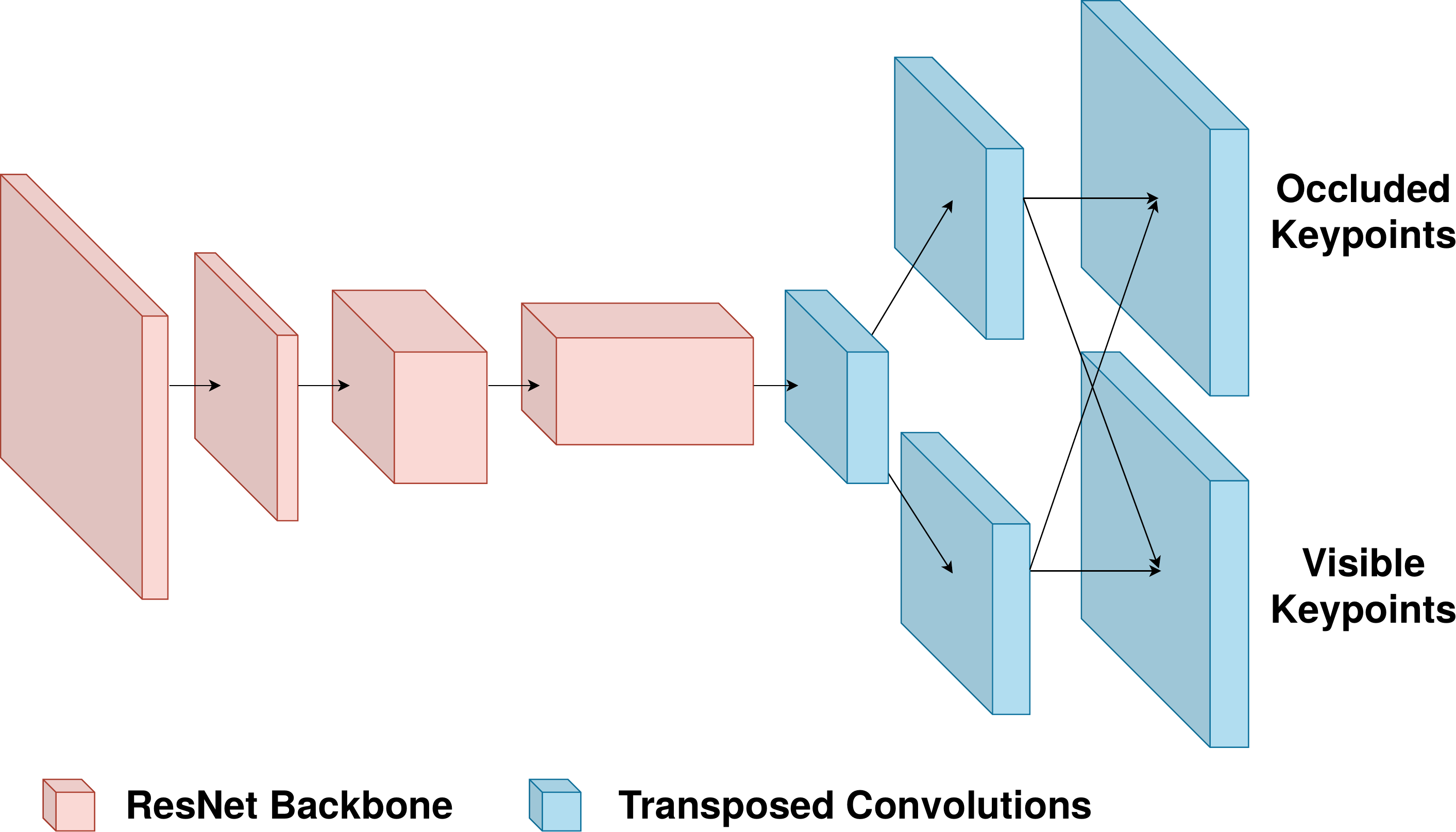}
    \end{center}
    \caption{\textbf{Our two architecture extensions: OccNet (top), OccNetCB (bottom)}. The Occlusion Detection Networks with an additional branch to detect occluded keypoints and OccNetCB with interconnections between the visible and occluded branch.}
    \label{fig:occNetCB}
\end{figure}

In order to utilize the additional information provided by the occlusion flags in the JTA dataset, the baseline architecture is extended by integrating a branch exclusively for the detection of occluded keypoints. 
The idea behind that is that the detection of occluded and visible keypoints are connected but different tasks. 
Furthermore, this should enable the network to learn a better representation of the human body structure that is more robust at handling occlusions and provide explicit additional occlusion information for possible subsequent methods in an image analysis pipeline. 
Two architectures are proposed with an additional branch to detect the occluded keypoints, see Fig.~\ref{fig:occNetCB}.\\
The first proposed architecture, OccNet (Occlusion Net), splits after two transposed convolutions so a joint representation can be learned in the previous layers. 
The second architecture OccNetCB (Occlusion Net Cross Branch) splits after only one transposed convolution, however the output from both layers are shared to one another. 
So that both branches can utilize information extracted by the other one. 
JTA also provides the distinction between self-occluded and occluded keypoints, during training self-occluded keypoints are treated in the same way as visible keypoints.\\
The Occlusion Detection Networks output two sets of heatmaps per pose, one for visible keypoints and the other one for occluded keypoints. 
The mean squared error loss of the baseline model of Xiao \etal~\cite{xiao2018simple} is adjusted by adding a second set of heatmaps for occluded keypoints and a weighting factor $\alpha$ to account for the lesser share of occluded keypoints in the training data. 
The proposed loss can be defined as:
\begin{equation}
    \begin{split}
    \mathcal{L}_{o}=\frac{ 1 }{n}\sum_{i=1}^n\parallel P^{\text{vis}}_{ i } - G^{\text{vis}}_{i}\parallel_{ 2 } + \enspace
    \alpha \parallel P^{\text{occ}}_{ i } - G^{\text{occ}}_{ i }\parallel_{ 2 }
    \end{split}
\end{equation}
where $G_i$ and $P_i$ denote the respective ground-truth and predicted heatmap of the corresponding keypoint $i$.
Due to its definition, the loss explicitly penalizes keypoints predicted by the wrong branch, which is achieved by setting the ground truth heatmaps for visible keypoints to zero in the occluded branch and vice versa.

\begin{figure} 
    \centering
    \subfloat[CrowdPose]{{\includegraphics[height=6.38em]{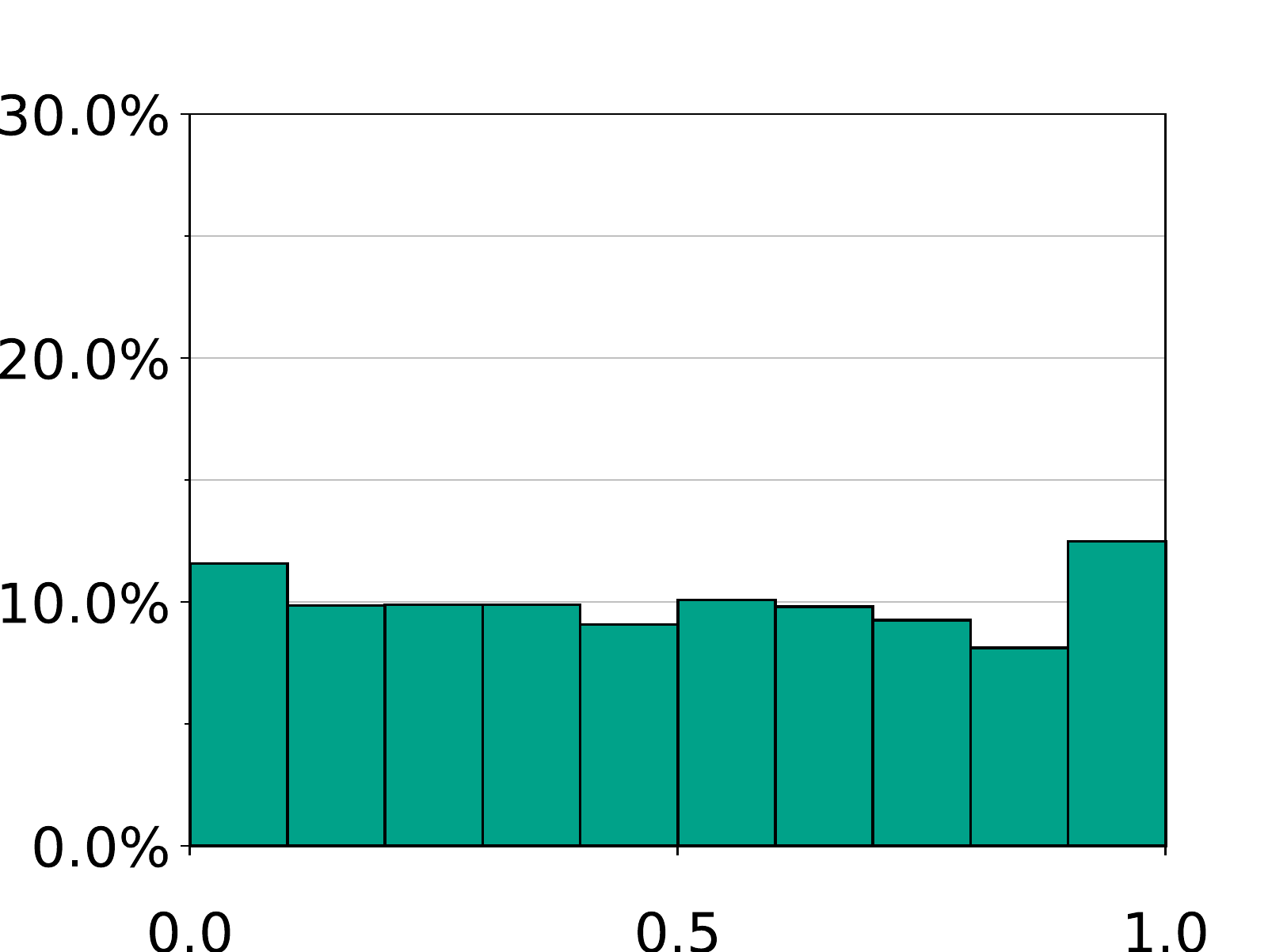} }}\hspace{-1.35em}%
    \subfloat[JTA]{{\includegraphics[height=6.38em]{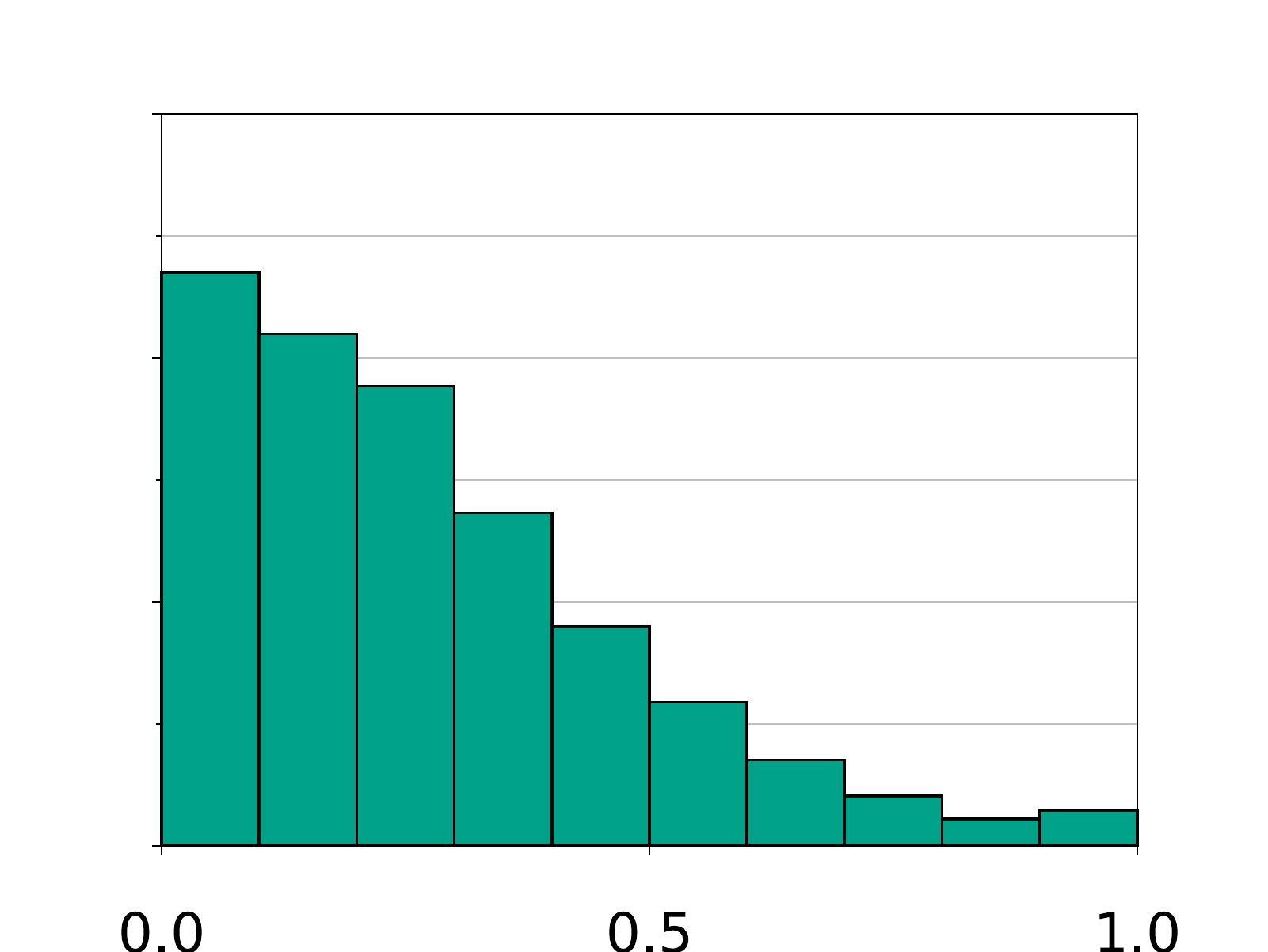} }}\hspace{-1.35em}%
    \subfloat[JTA-Ext]{{\includegraphics[height=6.38em]{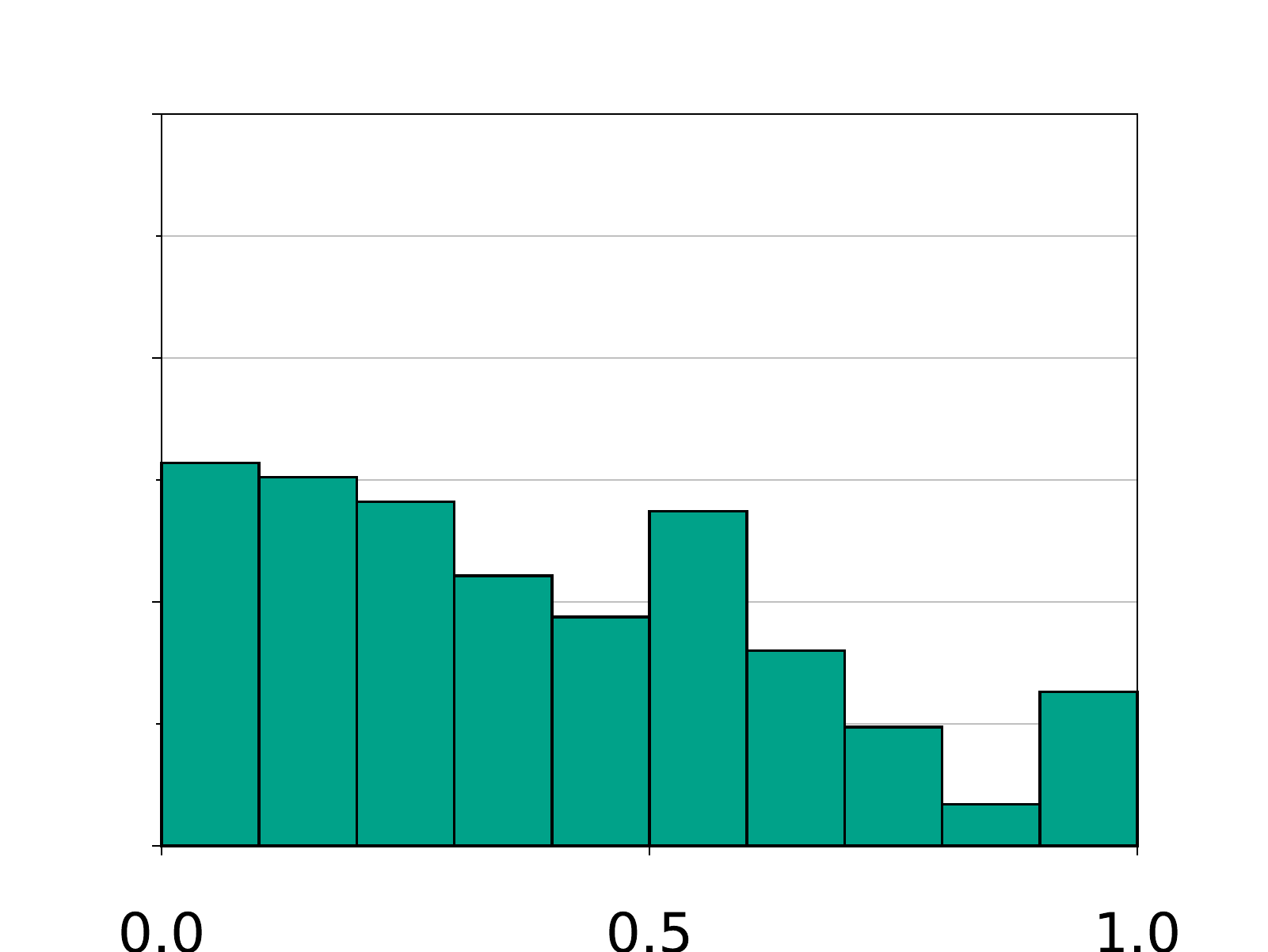} }}
    \caption{\textbf{CrowdIndex distributions}. CrowdPose and JTA differ significantly regarding their CrowdIndex distributions. JTA-Ext was created to diminish this difference and create a distribution closer to uniform distributed.}

    \label{fig:crowdindex}%
\end{figure}

\subsection{JTA-Ext}
\begin{figure*}
    \begin{center}
        \includegraphics[width=0.48\linewidth]{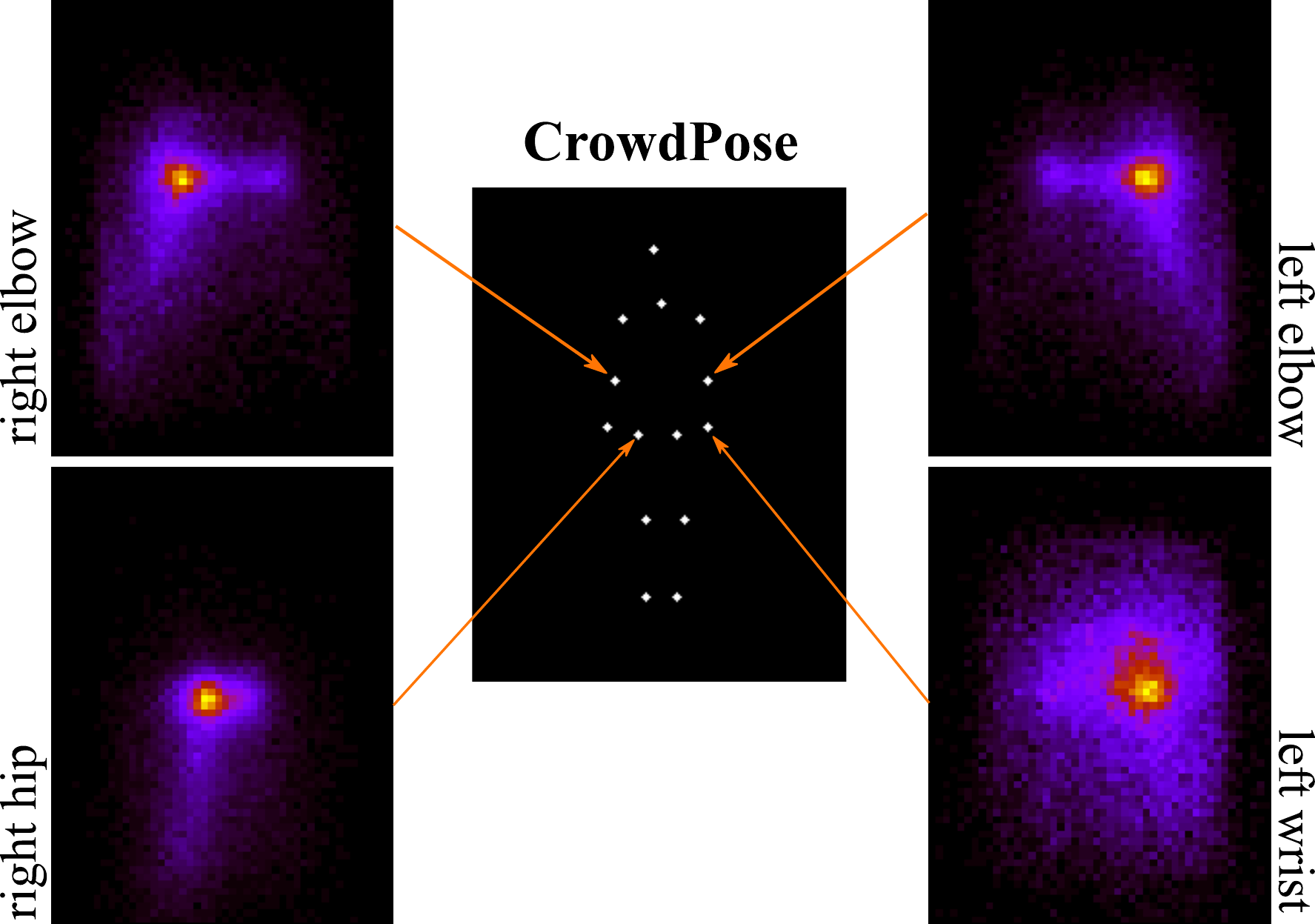}
        \hfill
        \includegraphics[width=0.48\linewidth]{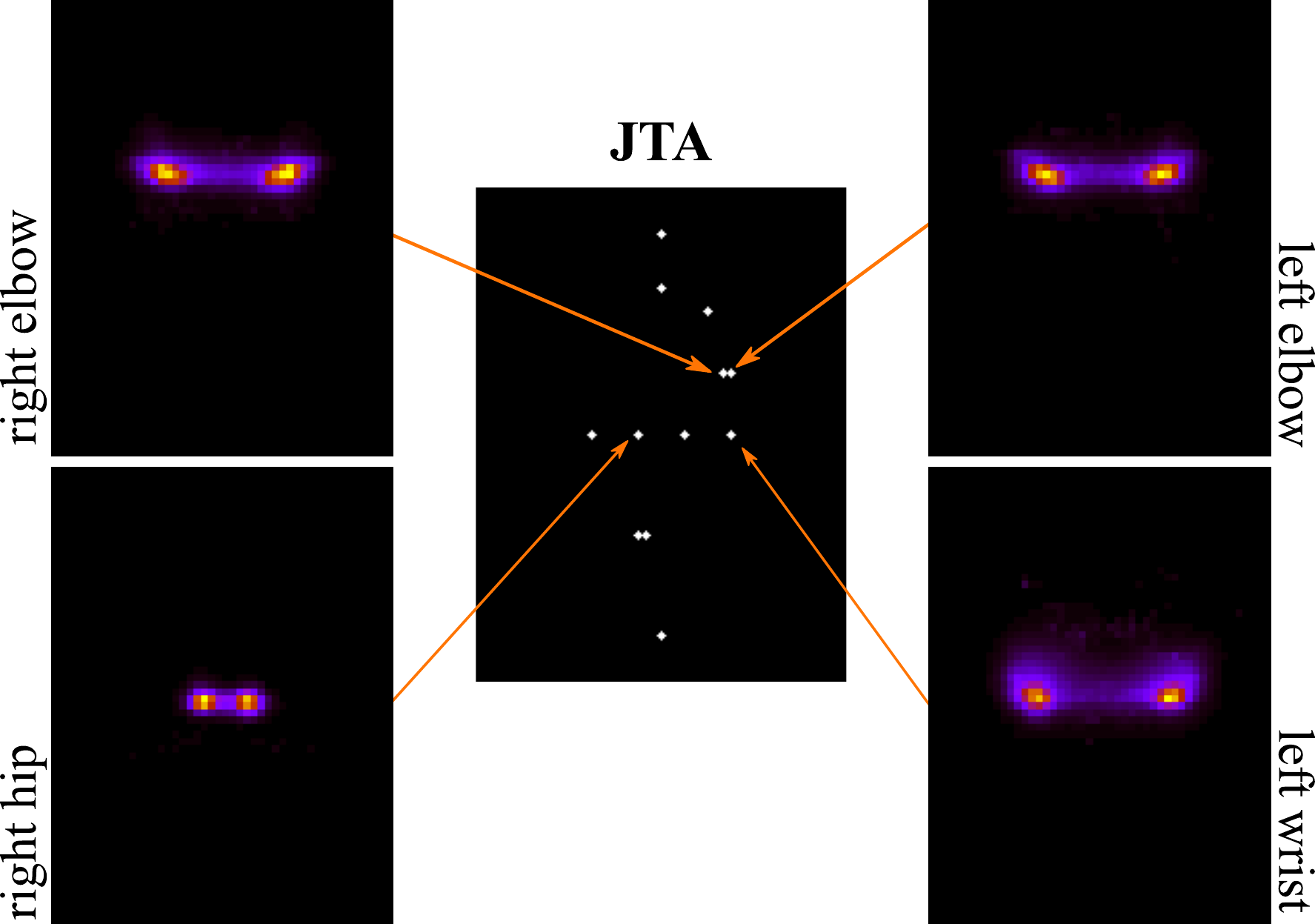}
    \end{center}
    \caption{\textbf{2d histograms of selected keypoint types of CrowdPose and JTA}. It is evident that the keypoint distribution of CrowdPose is more spread out than the distributions of JTA. However, the distributions of JTA are multivariate, because the persons in JTA face the camera just as often as they do not, whereas the persons in CrowdPose are primarily facing the camera. This is also represented in the average poses for the respective datasets.}
    \label{fig:avg_pose}
\end{figure*}

The synthetic dataset JTA introduced by Fabbri \etal~\cite{fabbri2018learning} is currently the largest publicly available pose tracking dataset with a larger number of people within an image. 
However, the adaption of the dataset for real-world crowd applications introduces some challenges, mainly: the visual difference of synthetic images of a video game, limited pose variety and less crowded scenes in comparison to CrowdPose.
Since JTA focuses primarily on urban settings, the images almost exclusively show people walking, which leads to far less variety of poses. 
In order to quantify this observation, Fig.~\ref{fig:avg_pose} shows 2d histograms of the location of specific keypoint types. 
It is evident that the distribution of the JTA keypoints are much more concentrated in comparison to the approximately uniformly distributed CrowdPose dataset, which leads to the problem a model purely trained on JTA has a strong bias of walking and standing people. 
We addressed this issue by extending the existing dataset and including all activities provided with the JTA-Mod like sitting, doing yoga, doing push-ups, cheering and fighting.
In this work, the proposed extended JTA version is referred to as \emph{JTA-Ext}\footnote{Dataset available on \url{https://github.com/thomasgolda/Human-Pose-Estimation-for-Real-World-Crowded-Scenarios}}.
Another goal in creating an extension for the JTA dataset was to increase the share of highly crowded scenarios. 
In order to gauge the "crowdedness" of the datasets we utilized the CrowdIndex proposed by Li~\etal~\cite{li2018crowdpose}.
The different CrowdIndex distributions are presented in Fig. \ref{fig:crowdindex}. 
While \cite{li2018crowdpose} sampled the dataset to achieve a uniform CrowdIndex distribution, JTA has a bias on less crowded images. 
The proposed extension is aimed to increases the amount of strongly crowded images in the~dataset.
The final extended JTA dataset consists of 58 additional video sequences with a total of 812,742 poses distributed over 46,350 frames.

\section{Experiments}\label{sec:experiments}

\subsection{Datasets} 
We conduct experiments on two different pose estimation datasets: the recently released real-world dataset CrowdPose~\cite{li2018crowdpose} and the synthetical dataset JTA~\cite{fabbri2018learning}.\\
The CrowdPose dataset \cite{li2018crowdpose} contains about 20,000 images and a total of 80,000 human poses with 14 labeled keypoints. 
In the following experiments the approaches will be trained on the training set of CrowdPose and will be evaluated on the test set of CrowdPose which includes 8,000~images.\\
JTA is a synthetic dataset for pedestrian pose estimation and tracking which Fabbri \etal~\cite{fabbri2018learning} collected by extracting videos from the video game \emph{Grand Theft Auto V}.
The annotations for human poses in JTA include 22 keypoints, a tracking id and visibility flags.
The visibility flags distinguish whether a keypoints is visible, occluded or self-occluded.
In order to compare the approaches across different datasets the annotations of JTA were matched to the format of CrowdPose by simply discarding and reordering the keypoints.
For the evaluation on the JTA dataset only a subset of the test set is used which includes only every fifteenth frame of the JTA test-sequences.
Furthermore, JTA-Ext is used for training which includes the extension of the JTA dataset and the original training set of JTA.

\begin{figure}[ht]
    \centering
    \subfloat[CrowdPose]{{\includegraphics[width=0.48\linewidth]{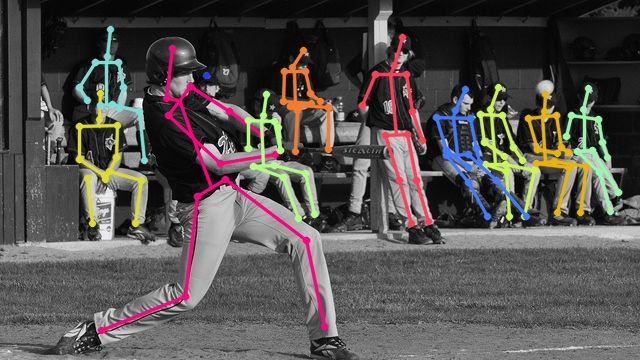}\label{subfig:crowded_crowdpose} }}%
    \subfloat[Surveillance]{{\includegraphics[width=0.48\linewidth]{images/006_pose.jpg}\label{subfig:crowded_crowdpe} }}%
    \caption{\textbf{Two images showing crowded situations}. Based on the CrowdIndex Fig.~\ref{subfig:crowded_crowdpose} belongs to the class of hard cases and Fig.~\ref{subfig:crowded_crowdpe} to the medium cases.}
    \label{fig:crowdedness}%
\end{figure}

\vspace{-.75em}

\subsection{Evaluation Metrics}
Following the procedure introduced in \cite{li2018crowdpose}, the adjusted Object Keypoints Similarity (OKS) of CrowdPose is used.
The OKS is used to describe the similarity of two poses and their keypoints by introducing a way to measure overlaps between keypoints in a manner as the IoU provides for bounding boxes.
Afterwards, the results are reported on three different partitions of the dataset which cluster images based on their CrowdIndex $C=\min\{\frac{1}{N} \sum_{i=1}^{N}\frac{N_a^i}{N_b^i},1.0\}$.
Thereby, $N$ is the number of persons in the image, $N_b^i$ is the number of keypoints within the bounding box of person $i$ corresponding to the person and $N_a^i$ is the number of keypoints inside bounding box belonging to all persons~$j \neq i$. 
These levels are denoted as easy $[0.0,0.1)$, medium $[0.1,0.8)$, and hard $[0.8,1.0]$.
Fig.~\ref{fig:crowdedness} shows two examples for crowded images.\\
Although the metric describes crowded images from a shallow viewing angle in an appropriate way, it does not really capture the crowdedness of images taken from a higher viewing point.
Due to this property, persons close to each other show a lower overlap of bounding boxes compared to image material recorded with lower located cameras.
Consequently, this results in a CrowdIndex of $0.91$ and $0.69$ for Fig.~\ref{subfig:crowded_crowdpose} and Fig.~\ref{subfig:crowded_crowdpe}~respectively, which suggests that the left image is more difficult.

\subsection{Implementation}
For the baseline model a TensorFlow re-implementation\footnote{\url{https://github.com/mks0601/TF-SimpleHumanPose}} of the model of Xiao \etal~\cite{xiao2018simple} is used with a pre-trained ResNet50 backbone. 
For the input images the same resolution of $256 \times 192$ is used while the resolution of the keypoint heatmaps is $64 \times 48$ as proposed by \cite{xiao2018simple}.
While ground-truth boxes are used during training of the models, the evaluation requires person detections for a fair comparison to other approaches. 
Therefore, the person detection results of the YOLOv3~\cite{DBLP:journals/corr/abs-1804-02767} model integrated in AlphaPose is used for all the following experiments\footnote{The code and the used weights of the used YOLOv3 provided by Xiu \etal~\cite{xiu2018poseflow} and Fang \etal~\cite{fang2017rmpe} are available at: \url{https://github.com/MVIG-SJTU/AlphaPose/tree/7eea793e5d497e7947bb8e6ce547d98c6a0059bf}} on CrowdPose. 
For our used loss function we set $\alpha$ to $1.5$.

\begin{table}[ht!]
	\begin{center}
	\caption{Results of the different proposed cutout augmentation methods on the CrowdPose test set. While all approaches improve the performance over all crowding levels, the object cutout augmentation provides the best result, with an increase in 1.1\% AP compared to the \baseline.}
	\label{tab:cutoutresCrowdpose}
	\vspace{5pt}
	\begin{tabular}{lcccc}
		\toprule
		\bfseries Cutout Method & \bfseries AP & \bfseries $\text{AP}_{\text{Easy}}$ & \bfseries $\text{AP}_{\text{Med}}$ & \bfseries $\text{AP}_{\text{Hard}}$ \\ 
		\midrule
        -                       & 61.9              & 72.7           & 63.1          & 49.1               \\ 
        \midrule
		 Objects                & \textbf{63.0 }    & \textbf{73.4} & \textbf{64.1} & \textbf{50.5}               \\ 
		 Body Parts             & 62.8              & 73.0          & 63.9          & 50.4               \\
		 Full Body              & 62.6              & 73.0          & 63.8          & 50.0    \\ 
		 \midrule
		 Parts \emph{and} Obj.  & 61.3              & 71.7          & 62.5          & 48.7    \\
		 Full  \emph{and} Obj.  & 61.5              & 71.6          & 62.7          & 49.0       \\ 
		 \midrule
		 Parts \emph{or} Obj.   & 62.9  	        & 73.0          & 64.0          & 50.2       \\
		 Full \emph{or} Obj.    & 62.8              & 73.1          & 63.9          & \textbf{50.5} \\ 
		 \bottomrule    
	\end{tabular}
	\end{center}
\end{table}

\subsection{Synthetic Generation of Occlusions}
For comparison between the different cutout methods and the effect these methods have in the crowded scenario the  \baseline model is trained on the CrowdPose training set and evaluated respective test set.
During training the Adam Optimizer is utilized with a learning rate of $10^{-3}$ which is dropped at epochs 90 and 120 by a factor of 10, similar to the training schedule proposed by Xiao \etal~\cite{xiao2018simple}. 
All the models were trained for a total number of 140 epochs with a mini-batch size of 64.
Additionally, to testing every single cutout method separately, combinations of objects and the different person cutout methods are evaluated. Therefore, two different setups are proposed. 
The first is, to add cutouts independently that means it is possible, that both cutout methods are applied simultaneously. The second one is, that the cutout methods are selected randomly with a probability of 50\% without both occurring at the same time.\\
The quantitative results in Table \ref{tab:cutoutresCrowdpose} show that all the proposed cutout methods improve the accuracy across all crowding levels, while the relative performance improvement increases as the scenes are more crowded.
The object cutout method produces the best results across all crowding levels on the CrowdPose test set, but the increase in performance in comparison to the body cutout methods stagnates as the images are more crowded. 
Another interesting observation is that the full body cutouts degrade the performance compared to adding only body parts. The reason behind that could be, that by avoiding the center of the image for the full body cutout, the degree of occlusion is lower than on the body part cutout, where the position is not limited. 
Applying two different cutout methods simultaneously leads to a decrease in accuracy. 
This can be attributed to the fact that the combination of two different cutouts within in image occlude too much of the person's body, prohibiting the network to see unobstructed people during training.  
Which would also explain that the accuracy increases again when only one of the two cutout method is applied per image. 

\begin{table}[hb!]
	\begin{center}
	\caption{Overview of the results of the occluded joint detection networks in comparison to the baseline architecture on the subset of the JTA test set. OccNetCB provides results similar to the \baseline model, whereas OccNet even degrades in performance.}
    \label{tab:occresultsjta}
	\begin{tabular}{lcccc}
		\toprule
		\bfseries Architecture  & \bfseries $\text{AP}$ & \bfseries $\text{AP}_{\text{Easy}}$ & \bfseries $\text{AP}_{\text{Med}}$ & \bfseries $\text{AP}_{\text{Hard}}$ \\ 
		\midrule
		\baseline              & \textbf{88.1}        & \textbf{90.8}               & \textbf{86.5}                 & \textbf{85.7}               \\
		OccNet                     & 87.8        & 90.7               & 86.3                 & 85.5               \\
		OccNetCB                    & 88.0        & 90.7               & \textbf{86.5}                 & \textbf{85.7} \\ 
		\bottomrule
	\end{tabular}
	\end{center}
\end{table}

\subsection{Detection of Occluded Keypoints}
In order to study the effects of explicitly detecting occluded keypoints in crowded pose estimation, the \baseline is compared to the proposed OccNet and OccNetCB architectures. Since the occluded joint detection networks need reliable ground truth annotations for visibility flags, the JTA dataset is chosen for training. 
All three architectures are trained for 15 epochs with a mini-batch size of 64 on the entire JTA training set. 
Again a learning rate of $10^{-3}$ is chosen which is dropped at epochs 8 and 10 by a factor of 10.
For testing a subset of the JTA test sequences is used. Also it is important to note that during evaluation the ground-truth human detections are used to ensure that occluded body parts are included in the boxes.\\
\begin{figure}[b!]
	\centering
	\includegraphics[width=0.24\linewidth]{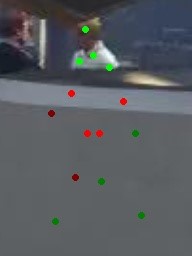}
	\includegraphics[width=0.24\linewidth]{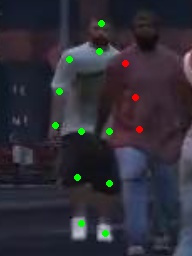}
	\includegraphics[width=0.24\linewidth]{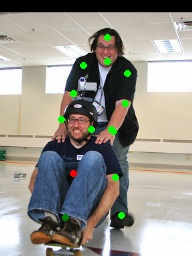}
	\includegraphics[width=0.24\linewidth]{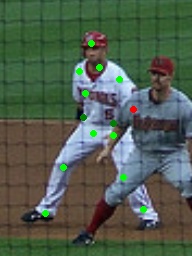}
	\caption{\textbf{Visualized results of the OccNetCB}. Detected occluded keypoints are marked in red and visible keypoints in green. The darker shade of keypoints denotes uncertain keypoints (i.e. with a likelihood lower than 0.7). The two images on the left show persons from JTA and the images on the right show selected images from CrowdPose \cite{li2018crowdpose}.}
	\label{fig:occdetections}
\end{figure}
The results in Table~\ref{tab:occresultsjta} show that the distinction between occluded and visible keypoints does not improve the accuracy, because the proposed methods achieve similar accuracy on the JTA test set. 
Nevertheless, the qualitative results presented in Fig.~\ref{fig:occdetections} show, that the proposed OccNet architectures learned to distinguish between visible and occluded keypoints, even on selected real-world images.
That proves, that the distinction between occluded and visible keypoints is working, but simply does not provide a benefit to pose estimation in crowded scenarios. 
Therefore, the results also imply that the \baseline model is sufficient to predict the location of occluded keypoints implicitly by inferring the location of visible keypoints.

\begin{table}[t]
	\begin{center}
	\caption{Comparison of the results of the \baseline model trained on JTA, JTA-Ext and CrowdPose, tested on the CrowdPose test set.}
    \label{tab:resultsJTA}
	\begin{tabular}{lcccc}
		\toprule
		\bfseries Training Set  & \bfseries $\text{AP}$ & \bfseries $\text{AP}_{\text{Easy}}$ & \bfseries $\text{AP}_{\text{Med}}$ & \bfseries $\text{AP}_{\text{Hard}}$ \\ 
		\midrule
		CrowdPose                    & 61.9        & 72.7               & 63.1                 & 49.1 \\ 
		\midrule
		JTA              & 26.6        & 36.6               & 27.0                 & 17.1               \\
		JTA-Ext                     & 28.8        & 39.8               & 29.2                 & 18.3               \\
		\bottomrule
	\end{tabular}
	\end{center}
\end{table}

\subsection{Synthetic Training Data}
Next, it is tested whether the use of synthetic data can provide comparable results on real-world datasets and if the extension of JTA can help to close the transfer-gap to a real-world dataset like CrowdPose. 
Therefore, the results of the \baseline model trained on JTA, JTA-Ext and CrowdPose are compared to each other on the CrowdPose test set. The results are presented in Table~\ref{tab:resultsJTA}. 
While it is apparent that the extension of JTA helped to improve the results, the accuracy severely lacks behind a model trained on CrowdPose. 
However, Table \ref{tab:resultsJTAFineTuned} shows that if the models are finetuned on CrowdPose is the AP increases across all crowding levels compared to a model purely trained on CrowdPose. 
Under these circumstances synthetic data has proven again to be a valuable addition to existing real-world training data, but should not be used as sole foundation for real-world applications.

\begin{table}[h]
	\begin{center}
	\caption{The results of the \baseline model pre-trained on JTA, JTA-Ext and fine-tuned on CrowdPose in comparison to a model purely trained on CrowdPose. The CrowdPose test set is used for evaluation.}
    \label{tab:resultsJTAFineTuned}
	\begin{tabular}{lcccc}
		\toprule
		\bfseries Pre-Training Set  & \bfseries $\text{AP}$ & \bfseries $\text{AP}_{\text{Easy}}$ & \bfseries $\text{AP}_{\text{Med}}$ & \bfseries $\text{AP}_{\text{Hard}}$ \\ 
		\midrule
		-                    & 61.9        & 72.7               & 63.1                 & 49.1 \\ 
		\midrule
		JTA              & \textbf{63.3}        & \textbf{73.7}               & \textbf{64.4}                & \textbf{50.6}               \\
		JTA-Ext                     & 63.0        & 73.1               & 64.3                 & 50.4               \\
		\bottomrule
		            
	\end{tabular}
	\end{center}
\end{table}

\begin{figure}[ht]
    \centering
    \subfloat{{\includegraphics[width=0.48\linewidth]{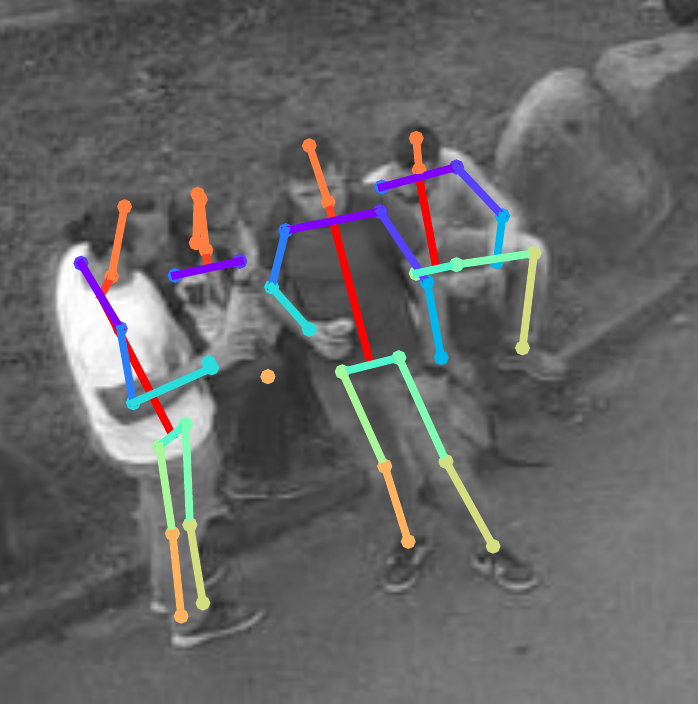}\label{subfig:res0b} }}%
    \subfloat{{\includegraphics[width=0.48\linewidth]{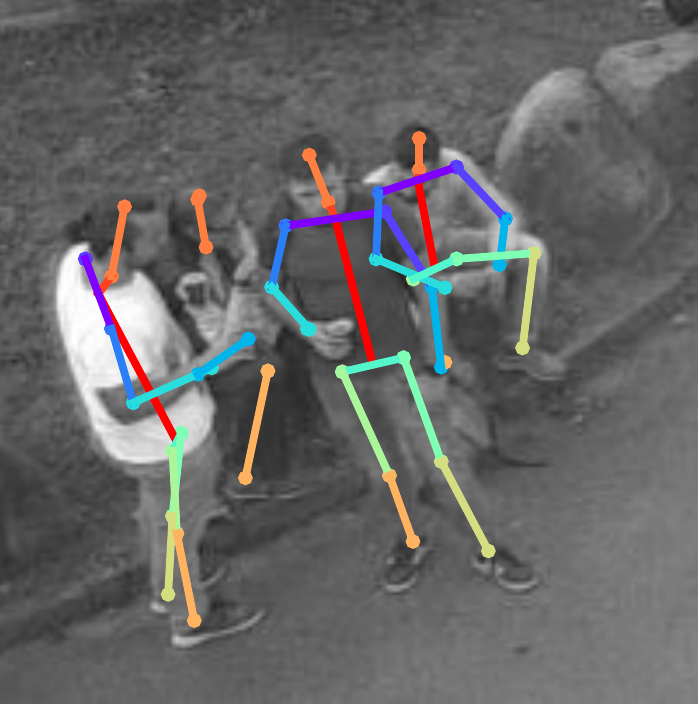}\label{subfig:res0f} }}%
    \\[-0.85em]
    \subfloat{{\includegraphics[width=0.48\linewidth]{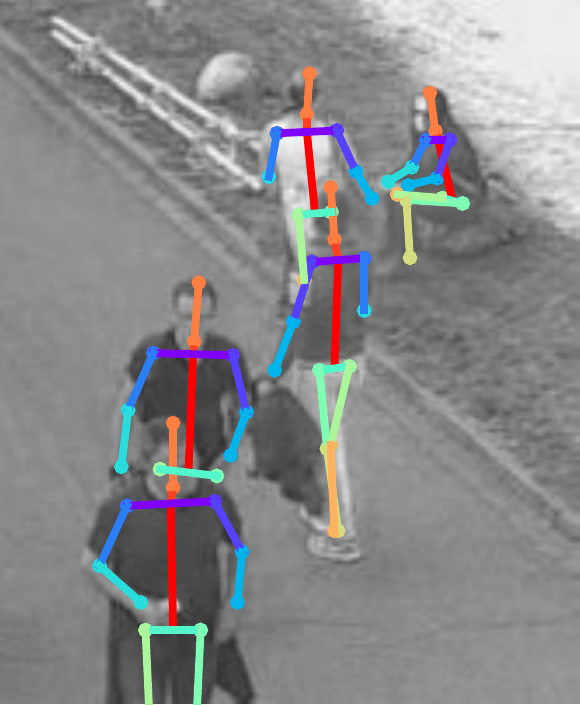}\label{subfig:res1b} }}%
    \subfloat{{\includegraphics[width=0.48\linewidth]{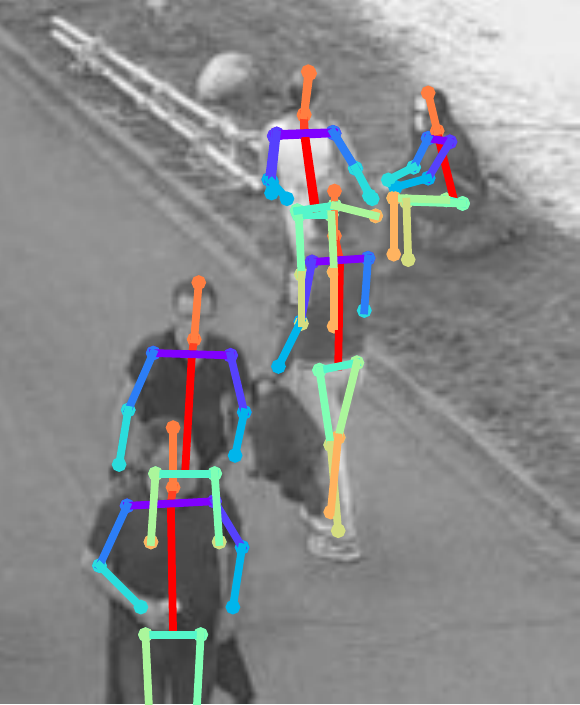}\label{subfig:res1f} }}%
    \\[-0.85em]
    \subfloat{{\includegraphics[width=0.48\linewidth]{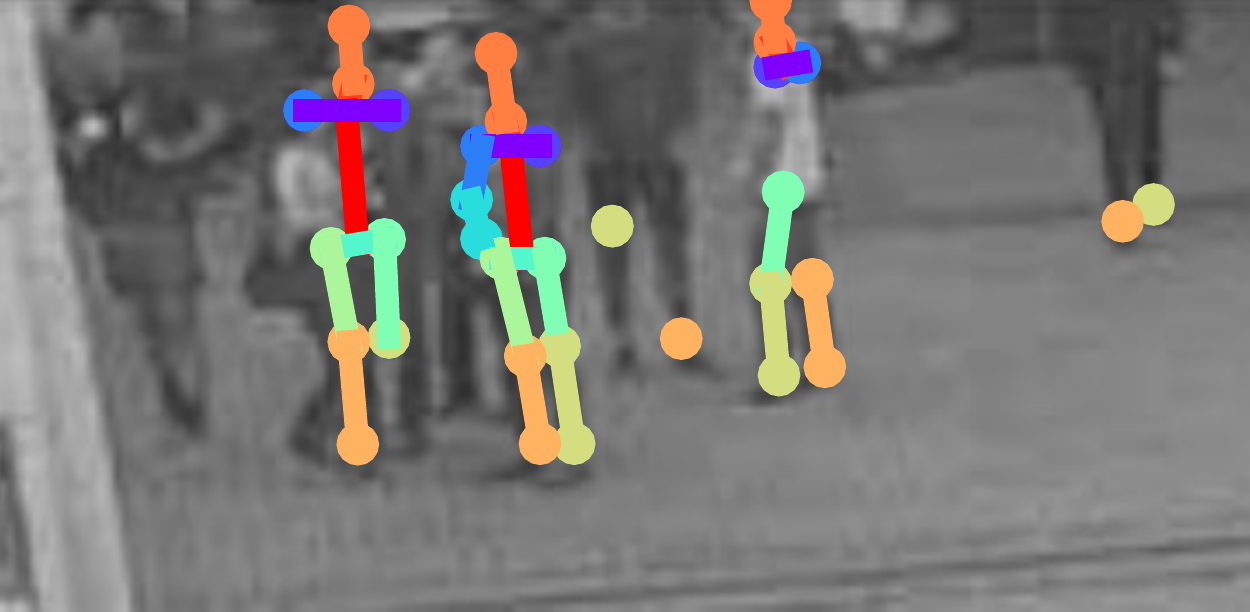}\label{subfig:res2b} }}%
    \subfloat{{\includegraphics[width=0.48\linewidth]{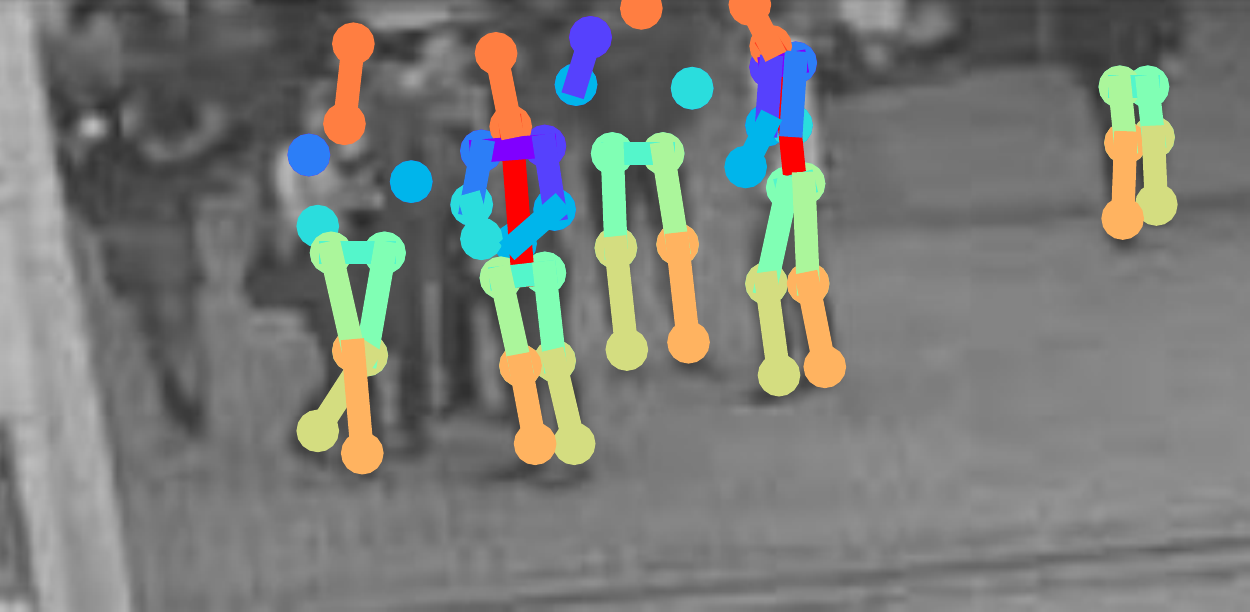}\label{subfig:res2f} }}%
    \caption{\textbf{Qualitative results}. The left column reports results generated by our \baseline method, the right column of our final model. The latter detects more keypoints and delivers legit estimates for occluded ones (see first and second row). Furthermore, at smaller scale more poses can be detected (third row). The crops have heights of 254px, 239px, and 97px for the top, middle and last row respectively.}
    \label{fig:qualitativeresults}%
\end{figure}

\subsection{Combined Experiments}
Previous experiments explored approaches to improve the baseline architecture and evaluated them separately for the task of crowded pose estimation. 
The following experiment combines these approaches to compete with current approaches on the CrowdPose dataset. 
In contrast to previous experiments a ResNet101 backbone is used to provide a fair comparison to the approach of \cite{xiao2018simple} and \cite{li2018crowdpose}. 
Additionally, the model is pre-trained on JTA for 8 epochs with a learning rate of $10^{-3}$ which is dropped at epochs 3 and 8 by a factor of 10. 
During training a mini-batch size of 80 was chosen and the object cutout method is utilized. 
For re-training on CrowdPose the same learning schedule and batch size as described in previous experiments was set. 
The results of Xiao \etal and Li \etal are taken from \cite{li2018crowdpose}, in which the models are purely trained on the CrowdPose training set. 
As stated above the same human detector of \cite{li2018crowdpose} is used to ensure a fair comparison between the different methods. 
\\
Our proposed approach shows significant improvement compared to the adapted baseline approach of \cite{xiao2018simple} and also provides comparable results to Li \etal~\cite{li2018crowdpose} on the easy and medium crowding levels. 
However, on extremely crowded images \cite{li2018crowdpose} surpasses the final model on CrowdPose by 4.3\% AP.
This can be attributed to their joint candidate matching method designed especially for handling these kinds of crowded scenarios.
However, to our best knowledge there are no further methods reporting results on CrowdPose.
A qualitative comparison of the improvements of our combined model in relation to the \baseline model is depicted in Fig.~\ref{fig:qualitativeresults}.

\begin{table}
	\begin{center}
	\caption{Results on the CrowdPose test set including the proposed split of \cite{li2018crowdpose} in different crowding levels. Since the dataset is quite young, only few work reports results on its test set.}
    \label{tab:resultsJTAFineTunedFinal}
	\begin{tabular}{lcccc}
		\toprule
		\bfseries Method  & \bfseries $\text{AP}$ & \bfseries $\text{AP}_{\text{Easy}}$ & \bfseries $\text{AP}_{\text{Med}}$ & \bfseries $\text{AP}_{\text{Hard}}$ \\
		\midrule
		Xiao \etal \cite{xiao2018simple}                    & 60.8        & 71.4               & 61.2                 & 51.2 \\ 
		Li \etal \cite{li2018crowdpose}              & \textbf{66.6}        & \textbf{75.7}               & 66.3                & \textbf{57.4}               \\ \midrule
		Ours                    & 65.5        & 75.2               & \textbf{66.6}                 & 53.1               \\
		\bottomrule
		            
	\end{tabular}
	\end{center}
\end{table}
\section{Conclusion}
In this work, we examined the impact of different extensions to the training of human pose estimators for crowded scenes.
We show that significant improvements were accomplished by utilizing simple data augmentation techniques like adding COCO~\cite{LinMBHPRDZ14} objects to the input images from crowded scenarios thus simulating occlusions. 
On the other hand, we observe that adding person instances or body parts is less effective and also requires additional attention to avoid creating erroneous poses in a top-down method. 
Our results further prove that JTA~\cite{fabbri2018learning} is a valuable addition to the training data for crowd-level pose estimation, especially since no large datasets are available for pose estimation in crowded scenarios. 
The created extension of JTA, which includes a higher variety of poses and denser crowds, further improves the accuracy on CrowdPose. 
However, solely relying on synthetic training data is not sufficient, due to the domain gap between synthetic and real-world data, originating from limited pose variety and visual limitations of a real-time rendering engine. 
Finally, the combination of the investigated approaches improved the baseline model and achieves results comparable to state-of-the-art methods on the real-world dataset CrowdPose.\\
In future work, we will especially focus on the problem arising by the gap between synthetic and real-world data.

{\small
\bibliographystyle{ieee}
\bibliography{egbib}

\begin{thebibliography}{10}\itemsep=-1pt

\bibitem{andriluka14cvpr}
M.~Andriluka, L.~Pishchulin, P.~Gehler, and B.~Schiele.
\newblock 2{D} {H}uman {P}ose {E}stimation: {N}ew {B}enchmark and {S}tate of
  the {A}rt {A}nalysis.
\newblock In {\em IEEE Conference on Computer Vision and Pattern Recognition
  (CVPR)}, June 2014.

\bibitem{pyramidnetwork}
Y.~Chen, Z.~Wang, Y.~Peng, Z.~Zhang, G.~Yu, and J.~Sun.
\newblock Cascaded pyramid network for multi-person pose estimation.
\newblock {\em CoRR}, abs/1711.07319, 2017.

\bibitem{fabbri2018learning}
M.~Fabbri, F.~Lanzi, S.~Calderara, A.~Palazzi, R.~Vezzani, and R.~Cucchiara.
\newblock Learning to detect and track visible and occluded body joints in a
  virtual world.
\newblock In {\em European Conference on Computer Vision (ECCV)}, 2018.

\bibitem{fang2017rmpe}
H.-S. Fang, S.~Xie, Y.-W. Tai, and C.~Lu.
\newblock {RMPE}: Regional multi-person pose estimation.
\newblock In {\em ICCV}, 2017.

\bibitem{7780459}
K.~{He}, X.~{Zhang}, S.~{Ren}, and J.~{Sun}.
\newblock Deep residual learning for image recognition.
\newblock In {\em 2016 IEEE Conference on Computer Vision and Pattern
  Recognition (CVPR)}, pages 770--778, June 2016.

\bibitem{li2018crowdpose}
J.~Li, C.~Wang, H.~Zhu, Y.~Mao, H.-S. Fang, and C.~Lu.
\newblock Crowdpose: Efficient crowded scenes pose estimation and a new
  benchmark.
\newblock {\em arXiv preprint arXiv:1812.00324}, 2018.

\bibitem{LinMBHPRDZ14}
T.~Lin, M.~Maire, S.~J. Belongie, L.~D. Bourdev, R.~B. Girshick, J.~Hays,
  P.~Perona, D.~Ramanan, P.~Doll{\'{a}}r, and C.~L. Zitnick.
\newblock Microsoft {COCO:} common objects in context.
\newblock {\em CoRR}, abs/1405.0312, 2014.

\bibitem{stackedhourglass}
A.~Newell, K.~Yang, and J.~Deng.
\newblock Stacked hourglass networks for human pose estimation.
\newblock {\em CoRR}, abs/1603.06937, 2016.

\bibitem{DBLP:journals/corr/abs-1804-02767}
J.~Redmon and A.~Farhadi.
\newblock Yolov3: An incremental improvement.
\newblock {\em CoRR}, abs/1804.02767, 2018.

\bibitem{Sarandi18IROSW}
I.~S{\'a}r{\'a}ndi, T.~Linder, K.~O. Arras, and B.~Leibe.
\newblock How robust is 3d human pose estimation to occlusion?
\newblock In {\em IROS Workshop - Robotic Co-workers 4.0}, 2018.

\bibitem{xiao2018simple}
B.~Xiao, H.~Wu, and Y.~Wei.
\newblock Simple baselines for human pose estimation and tracking.
\newblock In {\em European Conference on Computer Vision (ECCV)}, 2018.

\bibitem{xiu2018poseflow}
Y.~Xiu, J.~Li, H.~Wang, Y.~Fang, and C.~Lu.
\newblock {Pose Flow}: Efficient online pose tracking.
\newblock In {\em BMVC}, 2018.

\end{thebibliography}
}


\end{document}